\documentclass[letterpaper, 10 pt, conference]{ieeeconf}  

\IEEEoverridecommandlockouts                              

\usepackage{algorithm}
\usepackage{algpseudocode}
\usepackage{amsmath}
\usepackage{graphicx} 
\usepackage[skip=2pt]{caption}
\usepackage{afterpage}
\usepackage{xcolor}
\usepackage{comment}
\usepackage{float}

\title{\LARGE \bf
Meta-Adversarial Inverse Reinforcement Learning\\ for Decision-making Tasks
}

\author{Pin Wang$^{1}$ \and Hanhan Li$^{2}$ \and Ching-Yao Chan$^{1}$%
\thanks{$^{1}$ University of California, Berkeley
{\tt\small \{pin\_wang,cychan\} @berkeley.edu}}%
\thanks{$^{2}$ Google Research 
{\tt\small uniqueness@google.com}}%
}

\begin{document}

\maketitle
\thispagestyle{empty}
\pagestyle{empty}

\begin{abstract}
     Learning from demonstrations has made great progress over the past few years. However, it is generally data hungry and task specific. In other words, it requires a large amount of data to train a decent model on a particular task, and the model often fails to generalize to new tasks that have a different distribution. In practice, demonstrations from new tasks will be continuously observed and the data might be unlabeled or only partially labeled. Therefore, it is desirable for the trained model to adapt to new tasks that have limited data samples available. In this work, we build an adaptable imitation learning model based on the integration of Meta-learning and Adversarial Inverse Reinforcement Learning (Meta-AIRL). We exploit the adversarial learning and inverse reinforcement learning mechanisms to learn policies and reward functions simultaneously from available training tasks and then adapt them to new tasks with the meta-learning framework. Simulation results show that the adapted policy trained with Meta-AIRL can effectively learn from limited number of demonstrations, and quickly reach the performance comparable to that of the experts on unseen tasks.
    
\end{abstract}

\section{Introduction}
	
    Learning from demonstrations has gained a lot of popularity in decision-making and control tasks in a number of domains such as gaming, human-computer interactions, robotics, and self-driving vehicles. Commonly used methods include Imitation Learning (IL)~\cite{bojarski2016end}~\cite{ross2011reduction}, Inverse Reinforcement Learning (IRL)~\cite{ng2000algorithms}~\cite{ziebart2008maximum}, Generative Adversarial Imitation Learning (GAIL)~\cite{ho2016generative}, Adversarial Inverse Reinforcement Learning (AIRL)~\cite{fu2017learning}~\cite{wang2019human}, etc. These methods do not require designing a reward function manually that can be hard for complex tasks, but they generally do need an abundance of demonstrations in order to mimick the demonstrated behaviors. Furthermore, the learned behavior usually works only in that specific task environment and fails to generalize to new tasks with different distributions. In reality, it is usually the case that we continuously enrich the dataset by collecting data from new tasks or environments. For example, to learn an automated lane-change behavior, we may train our vehicle agent with thousands or even millions of labeled driving demonstrations from different cities or countries, but these demonstrations may not cover all the possible situations and we may still have new data obtained from other locations that are not originally included in our training dataset. In this situation, it is labor intensive and costly to keep labeling all the newly acquired data and retrain the model from scratch again. Therefore, it is essential to have a model that can make good use of the knowledge learned from existing tasks and generalize quickly to new tasks with limited data samples. 
    
    There have been a number of studies that applied meta-learning to decision-making and control tasks, such as Meta-Reinforcement Learning~\cite{wang2016learning}~\cite{xu2018meta}, Meta-Inverse Reinforcement Learning~\cite{xu2018learning}~\cite{yu2019meta}, Meta-DAgger~\cite{sallab2017meta}, etc. These studies target on either policy learning or reward function learning. For a decision-making and control task, what we ultimately prefer is a policy that maps states to actions for maneuvering the agent to complete a task, but we may still want a reward function that is considered as a succinct, robust and transferable representation of a task~\cite{abbeel2004apprenticeship}. Some of the learning from demonstrations methods, e.g. Adversarial Inverse Reinforcement Learning (AIRL)~\cite{fu2017learning}, can simultaneously recover a policy and a reward function from expert demonstrations for single task. In this work, we propose a framework to combine Meta-learning and Adversarial Inverse Reinforcement Learning to learn the model initialization that can be quickly adapted to new situations with limited data. 
    
    Our contributions in this work are as follows: 1) We proposed a general Meta-AIRL algorithm that can efficiently generalize both of the learned policy and reward function to new tasks. 2) We applied the proposed method to the decision-making tasks in autonomous driving domain with the consideration that such an application is challenging as the learning agent needs to dynamically interact with adjacent agents to make proper actions.

    The rest of the paper is organized as follows. In Section 2, we present related work. Section 3 provides details on the formulation of the Meta-AIRL methodology. Experiments and results are given in Section 4. Section 5 concludes this paper.
    

\section{Related Work}
\label{sec:citations}

\subsection{Learning from Demonstrations}
    
    There has been a number of studies focusing on learning from demonstrations.Pure Behavior Cloning~\cite{bojarski2016end} suffers from the dilemma of distribution shift that leads to the failure of issuing correct actions when the states are out of the training distribution. DAgger~\cite{ross2011reduction} solves the problem by introducing a human expert or an oracle to give corrective labels for some states that are not included in the collected data. However, human intervention in deployment is troublesome in many practical settings. Generative Adversarial Imitation Learning (GAIL)~\cite{ho2016generative} is proposed to learn a policy in an adversarial way by involving a discriminator and a generator (i.e. the policy). It saves the effort of inquiring a human annotation but uses a discriminator for distinguishing the expert data and the generated data. A limitation is that GAIL does not recover a reward function which is generally considered robust over different environments. Adversarial Inverse Reinforcement Learning (AIRL)~\cite{fu2017learning} formulates the discriminator in a special form instead of directly using neural network output, which enables AIRL to learn both the policy and reward function simultaneously.
    
    
\subsection{Meta-learning}
    
    Meta-learning, also known as learning to learn, is an approach to adapt learned models to novel settings by exploiting the inherent structural similarities across a distribution of tasks. Meta-learning can be metrics based~\cite{vinyals2016matching}~\cite{sung2018learning},  model based ~\cite{ravi2016optimization}, or optimization based ~\cite{ravi2016optimization}~\cite{finn2017model}~\cite{nichol2018first}. Optimization based meta-learning is a powerful approach that adjusts the optimization algorithm itself for fast learning. For example, Model-agnostic Meta-learning (MAML)~\cite{finn2017model} learns a good parameter initialization by performing one-step unrolling only. First-order MAML (FOMAML) and REPTILE~\cite{nichol2018first} algorithms learns the initialization by looking ahead multiple gradient steps but ignores second order derivatives. The learnt initialization by these methods encode common knowledge among different tasks and allows efficient adaptation in only a few gradient steps.

\subsection{Integration of Meta-learning and Learning from Demonstrations}
    
    Some studies have explored integrating Meta-learning and learning from demonstrations methods for decision-making and control tasks. One study is Meta-IRL, which combines Meta-learning and IRL to adapt the learned reward function to new tasks. In the studies of ~\cite{xu2018learning}~\cite{gleave2018multi}, Meta-IRL has been applied to solve problems with discrete tabular settings, while in the study of~\cite{yu2019meta}, the authors extend the standard IRL framework to include latent variables and learn an adaptable reward function in an unsupervised way. The approach has been applied to continuous control tasks in Mujoco environments~\cite{todorov2012mujoco}. Xu. et al.~\cite{xu2018learning} and Yu. et al.~\cite{yu2019meta} leveraged Meta-IRL to learn prior distributions of the tasks. These approaches have been generally applied to simulated game environments where the background is static and no interactions between agents are involved. It can be intractable to model the distribution of tasks for highly dynamic environments such as a driving domain.
    
    
    Autonomous driving is a much more challenging task in terms of the dynamically changing environment and complex interactions with surrounding agents. Recently, a couple of studies have explored to apply meta-learning in this domain. \cite{sallab2017meta} proposed Meta-DAgger to improve the generalization of the learned policy in a simulated environment called TORCS. It involves a human or an oracle in the loop which limits the application in the real world. \cite{jaafra2019context} is another work that integrated a policy iteration based RL into the meta-learning pipeline, and was tested for lane-keeping task in a simulated environment in CARLA. It requires the access to a known reward function. 
    Our work aims to learn both a policy and a reward function that can be effectively adapted to new tasks by learning from limited demonstrations, and we target on more complicated decision-making problems under dynamic environments where the learning agent intensively interacts with its surrounding agents.


\section{Meta-Adversarial Inverse Reinforcement Learning}
    
    
\subsection{Preliminaries}
\subsubsection{Adversarial Inverse Reinforcement Learning}
\label{AIRL}
    
    Adversarial Inverse Reinforcement Learning~\cite{fu2017learning} builds on Adversarial Learning and  maximum causal entropy Inverse Reinforcement Learning framework (IRL)~\cite{ziebart2010modeling}. 
    It introduces a discriminator to obtain the formulation of the reward function. The optimization in AIRL is similar to that in GAN~\cite{radford2015unsupervised} where the discriminator and generator are updated in an adversarial way. The difference is that AIRL takes a special form of the discriminator $D_{\omega}(s,a)$ instead of directly using the output of the neural network $f_{\omega}(s,a)$ as the discriminator value. The discriminator is given as:
    \begin{equation} 
    \label{disc}
        D_{\omega}(s,a) = \frac{\exp(f_{\omega}(s,a))}{\exp(f_{\omega}(s,a))+\pi(a|s)}
    \end{equation}
    where $s$ is the state, $a$ is the action, $\omega$ is the discriminator's parameter, and $\pi$ is the updated policy. 

\subsubsection{Meta Learning}
   
    Meta-learning involves a meta-learner and a learner. The learner learns the model $f_{\theta_{T_i}}$ from individual tasks $T_i$ in an inner loop, while the meta-learner trains the learner to get an overall common structure $f_{\theta}$ over a set of training tasks $T=\{T_i|i=1,...,N\}$ in an outer loop. 
    
    In our study, we leverage REPTILE pipeline as it does not unroll a computation graph or calculate any second derivatives. This feature facilitates the convenience of performing multiple gradient steps on the training tasks in the inner loop. More importantly, it allows us to apply different update frequency for the discriminator and the generator. 
    
    
   
    
    

\subsection{Meta-Adversarial Inverse Reinforcement Learning} 
\label{method}
\subsubsection{Loss Definition}
    
    
    In our study, we use the cross-entropy loss $L_D(\omega)$ in Equation (\ref{D_loss}) as the objective of the discriminator that performs a  binary classification task. We label the expert data as $1$ and the generated data as $0$.  Instead of using the full trajectory as input to the discriminator which might result in high variance, we input single state-action pairs $(s, a)$ at individual time steps to the discriminator, which has been proved more stable in learning~\cite{fu2017learning}.
    \begin{multline}
    \label{D_loss}
        L_{D}(\omega) = E_{(s,a)\sim \pi_{E}} [-\log D_{\omega}(s,a)] \\
        + E_{(s,a)\sim \pi_{\phi}} [-\log (1-D_{\omega}(s,a))]
    \end{multline}
    
    The generator represents the action policy of the agent whose optimization is based on a reward function. In our formulation, the reward function in Equation (\ref{reward}) is built upon the outcome of the discriminator and is considered as the connection between the discriminator and generator. The functionality of the reward function is to score high values for the generated data that confused the discriminator. 
    \begin{equation}
    \label{reward}
        r(s,a) = \log(D_{\omega}(s,a)) - \log (1-D_{\omega}(s,a))
    \end{equation}
    
    The objective of the generator is to generate trajectories as much similar as the expert demonstrations. That is achieved by maximizing the total rewards accumulated in an episode or up to a horizon $H$ in the policy optimization procedure. Actually, such a policy optimized is equivalent to an entropy regularized policy. 
    \begin{align}
    \label{E_pi}
    -L_G(\phi) =& E_{\pi_{\phi}} [\sum_{t=0}^{H} (\log (D_{\omega}(s_t,a_t)) - \log(1 - D_{\omega}(s_t,a_t)))] \nonumber\\
    =& E_{\pi_{\phi}} [\sum_{t=0}^{H} (f_{\omega}(s_t,a_t) - \log \pi(a_t|s_t))]
    \end{align}

\subsubsection{Meta Optimization}
    
    Within meta-training, we sample $N$ training tasks $\{T_i| i=1,...,N\}$ from the distribution $p(T)$. Each decision-making task include $K$ expert demonstrations, $S_{T_i}=\{\tau_1, ..., \tau_K\}$, where each demonstration $\tau_k$ is a sequence of state and action pairs $(s,a)_t, t\in\{1,...,H\}$. 
    
    In our study, the inner loop neural network in the meta-learning structure consists of two components, the discriminator and the generator, i.e., the parameters in the meta-learning model includes parameters from the two individual neural networks $\theta = \{\omega, \phi \}$. The loss of the overall model also includes two parts, $L = \{L_D(\omega), L_G(\phi) \}$. For the update in the inner loop, we use ADAM optimizer for the discriminator and Trust Region Policy Optimization (TRPO) for the generator, which are consistent with the AIRL and GAIL methods. 
    \begin{align}
        \frac{\partial L_{D}(\omega)}{\partial \omega} =& E_{(s,a)\sim \pi_{E}} [- \nabla_{\omega} \log (D_{\omega}(s,a))] \nonumber \\
        & + E_{(s,a)\sim \pi_{\phi}} [- \nabla_{\omega} \log (1-D_{\omega}(s,a))]     \label{grad_D}
    \end{align}
    \begin{align}
        \frac{\partial L_{G}(\phi)}{\partial \phi} = E_{(s,a)\sim \pi_{\phi}} [&\nabla_{\phi} \log \pi_{\phi}(a|s) (\log D_{\omega}(s,a) \nonumber\\
        &- \log (1-D_{\omega}(s,a)))]     \label{grad_G}
    \end{align}
    
    Note that the discriminator is used to provide reward signals to the generator, so it will be beneficial to have a good discriminator as soon as possible. Therefore, we use high update frequency $k_D$ for the discriminator and low frequency $k_G$ for the generator.
    
    In the outer loop, the model weights update is as follows: 
    \begin{equation}
    \label{meta_update}
        \omega \leftarrow \omega + \beta_\omega \frac{1}{N} \sum_{i=1}^N (\omega_{T_i} - \omega),   \phi \leftarrow \phi + \beta_\phi \frac{1}{N} \sum_{i=1}^N (\phi_{T_i} - \phi) 
    \end{equation}
    where $\omega_{T_i}$ and $\phi_{T_i}$ are the updated weights for the discriminator and the generator in the task $T_i$, $\beta = \{\beta_\omega, \beta_\phi\}$ are the meta learning rates. Similar to update frequencies, we set $\beta_\omega > \beta_\phi$, which gives better training stability.
    
    In our study, the data used for the discriminator update consists of two sets, an expert dataset $\mathcal{S}^E$ that is collected from simulated expert demonstrations and a generated dataset $\mathcal{S}^G$ that is from the generator under the current policy $\pi_{\phi}$. Each episode, $\tau = \{(s_1, a_1), ..., (s_H, a_H)\}$, includes a sequence of state-action pairs of $H$ steps that might vary in different episodes in our study. The overall algorithm is given in Algorithm \ref{algo}.
    
    \begin{algorithm}[t!]
    \caption{Meta Adversarial IRL}
    \label{algo}
    \begin{algorithmic}[1]
        \State Initialize $\theta=\{\omega,\phi\}$, $\omega$: parameters of the discriminator, $\phi$: parameters of the policy
        \For {iteration = $1,..., M$} 
            \State Sample training tasks $T_i\sim p(T)$, $i\in \{1, 2, ..., N\}$
            \For{all tasks $T_i$}
                \State Obtain expert demos $\mathcal{S}_{T_i}^D$
                \For{update = $1,2,...,K$}
                    \State Obtain generated demos $\mathcal{S}_{T_i}^G$ under $\pi_{\phi_{T_i}}$
                    \State Calculate $L_{D}(\omega_{T_i})$ with $\mathcal{S}_{T_i}^D$ and $\mathcal{S}_{T_i}^G$ 
                    \State Update discriminator parameters $\omega_{T_i} \leftarrow \omega_{T_i}-\alpha_D \nabla_{\omega} L_{D}(\omega_{T_i})$ for $k_D$ iterations
                    \State Calculate reward $r_{T_i}$ with $\mathcal{S}_{T_i}^G$ and $D_{\omega_{T_i}}$
                    \State Calculate $L_{G}(\phi_{T_i})$ with $\mathcal{S}_{T_i}^G$ and $r_{T_i}$
                    \State Update policy parameters $\phi_{T_i} \leftarrow \phi_{T_i}-\alpha_{G} \nabla_{\phi} L_{G}(\phi_{T_i})$ for $k_G$ iterations
                \EndFor
                \State Obtain parameters $\theta_{T_i}=\{\omega_{T_i}, \phi_{T_i}\}$
            \EndFor
            \State Update $\theta \leftarrow \theta + \beta \frac{1}{N} \sum_{i=1}^N (\theta_{T_i} - \theta)$
        \EndFor
    \end{algorithmic}
    \end{algorithm}
    
\section{Experiments}
\subsection{Application Case}

    We test our proposed method on decision-making tasks for lane-change scenarios in autonomous driving. The task is to decide when and where to make a lane-change maneuver. For example, when a vehicle receives a lane-change command from a route navigation module, our proposed method is to tell the vehicle at what time and to which vehicle gap it should make the lane change maneuver. 
    
    It is common that the expert demonstrations are collected by drivers with different personalities and in different regions. Particularly, in terms of the decision-making task, different driving styles show different preferences of vehicle merging gaps, safety distance, maximum/minimum accelerations, etc., resulting in different data distributions in the states and actions. The driving tasks in our study are considered drawn from such a distribution of different driving styles. As most off-the-shelf simulators have not provided the functionality of conveniently simulating diverse driving behaviors, we develop a new platform with oracle modules to simulate expert driving in multiple styles. The demonstrations include various behaviors, such as yielding behavior where the ego vehicle waits for a safety gap to merge in, overtaking behavior where the ego vehicle accelerates to merge to a gap in front of it on the adjacent lane, and aborting behavior where the ego vehicle aborts the lane changing when it detects potential collisions measured by a safety margin. In this study, we take the conservative and neutral driving styles as the meta-training tasks and the challenging style (aggressive driving) as the meta-testing task to check whether the generalized model really learns the encoded structure across tasks or just performs an average over tasks. 
    
\begin{figure}
        \centering
        \includegraphics[width=0.6\linewidth]{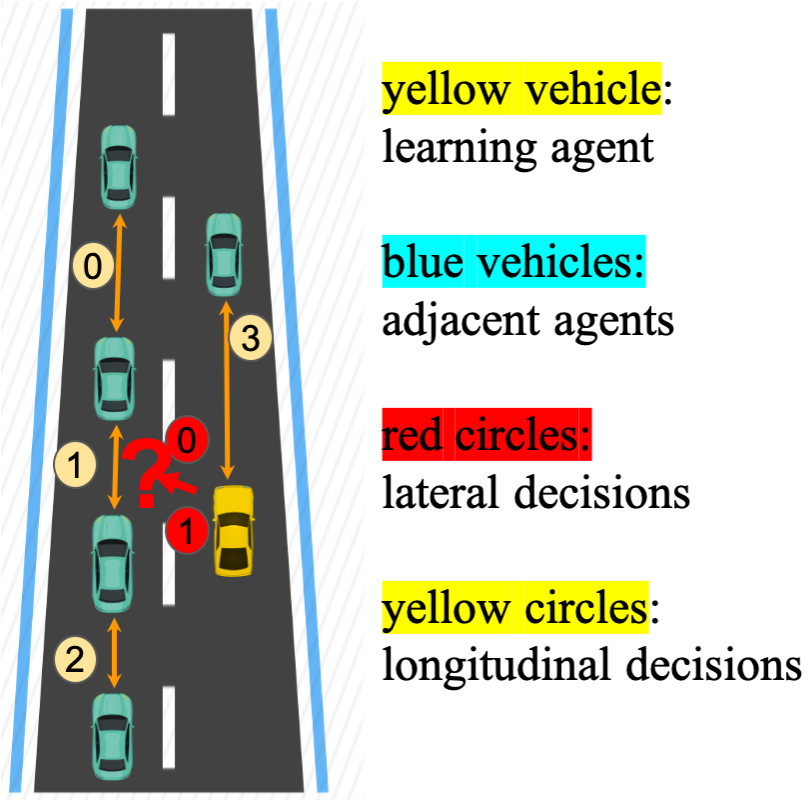}
        \setlength{\belowcaptionskip}{-10pt}
        \caption{\small An illustration of a simulated scenario. The learning agent (in yellow) interacts with its adjacent vehicles (in blue) and decides which vehicle gap (in yellow) to merge into and whether to make the lane change right now (in red).}
        \label{fig:simulation}
    \end{figure}

    As decisions are generally considered as discrete, we design the action space as discrete by joining the decisions on the lateral and longitudinal directions. Specifically, the longitudinal decision decides which vehicle gap to merge into and the lateral decision decides whether to make the lateral movement right now. An illustration of the simulated scenarios is shown in Figure~\ref{fig:simulation}. 
    
    The state space incorporates vehicle kinematic information of the ego vehicle and its surrounding vehicles. It includes information such as vehicle positions, speeds, accelerations, lane ids, etc., which constitute a 44-dimension state space. More experiment details are given in the next subsection. 
    
\subsection{Experiment Details}
    For the two meta-training tasks, i.e. conservative and neutral driving styles, each has 3000 demonstrations available for training, while the meta-testing task, i.e. aggressive driving task, only has access to a limited number of demonstrations such as $\{5, 10, .., 50\}$. 
    
    As mentioned in \ref{method}, the discriminator is trained with higher update frequency than the generator. We fine-tune this hyperparameter with the meta-training tasks and find the optimal numbers of iterations are $k_D=50$ for the discriminator and $k_G=1$ for the generator respectively. Also, for the meta learning rate in the outer loop, we fine-tune it and use $\beta_{\omega}=0.5$ for the discriminator and $\beta_{\phi}=0.25$ for the generator. The meta-training process is conducted for 2500 iterations. 
    In meta-testing, we test the generalization performance with different number of available demonstrations, i.e. $\{5,10,...,50\}$. For each case, we conduct 10 update iterations upon the model learned from meta-training. 
    
    


    \begin{figure*}[h]
        \centering
        \includegraphics[width=0.7\linewidth]{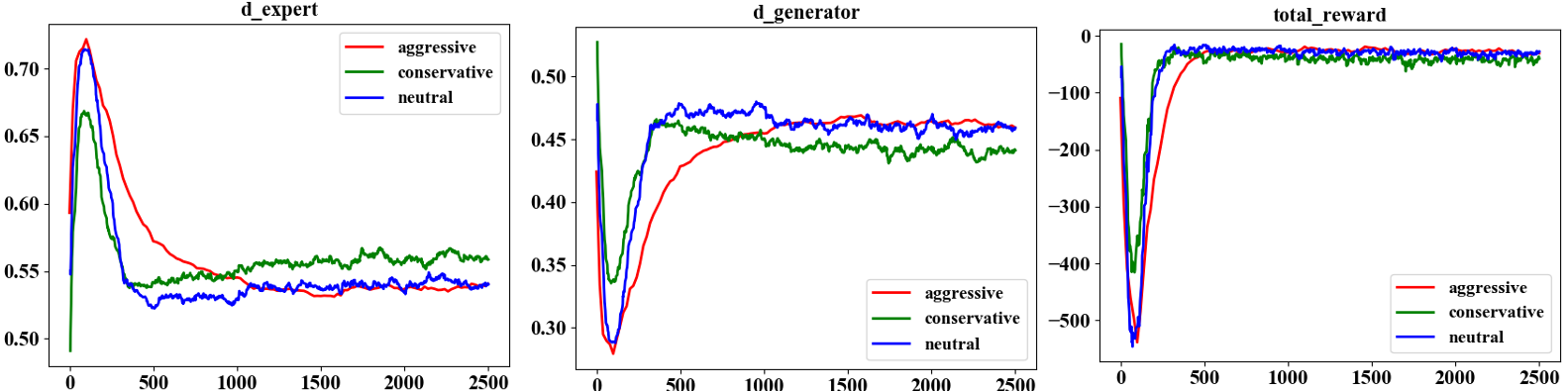}
        \setlength{\belowcaptionskip}{-10pt}
        \caption{\small Performance of the discriminator (left and middle graphs) and the generator (right graph). The left and middle graphs display the classification probabilities of the expert data and the generated data respectively from the discriminator. The high values in the experts and low values in the generator in early episodes (x-axis) indicate that the discriminator is able to distinguish the experts from the generator well. The values then go to around 0.5 at convergence for both the expert data and generated data as desired due to the adversarial effect with the generator. The right graph show the accumulated total rewards over training episodes. The increasing and plateauing trend  indicates the improved learning ability of the generator. (Blue and green curves represent the meta-training tasks, and the red curve represents the online-testing results of the meta-testing task.)}
        \label{fig:d_rwd}
    \end{figure*}

     \begin{figure*}[h]
        \centering
        \includegraphics[width=0.7\linewidth]{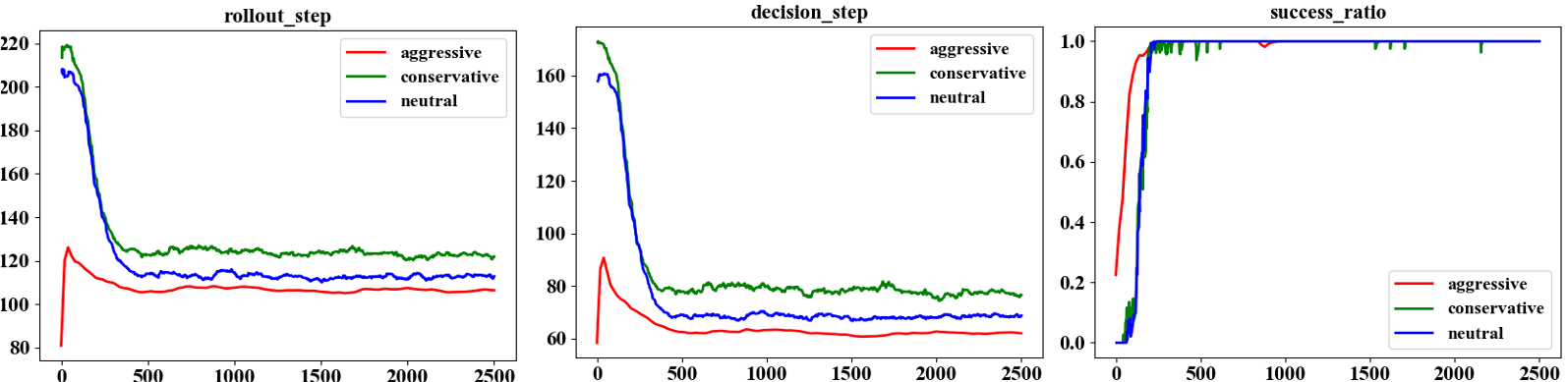}
        \setlength{\belowcaptionskip}{-10pt}
        \caption{\small Driving performance. The left and right graphs depict the agent's maneuvering performance in training via the roll-out steps and decision-making steps. The agents takes shorter and more stable time to make lane changes after trained for hundreds of episodes. The right graph displays the ratio curves of successful lane-change episodes over all experimenting episodes. The values are close to 1 after convergence, indicating that the agent can always make successful lane changes when requested.}
        \label{fig:steps_ratio}
    \end{figure*}

    \begin{figure*}[h]
        \centering
        \includegraphics[width=0.7\linewidth]{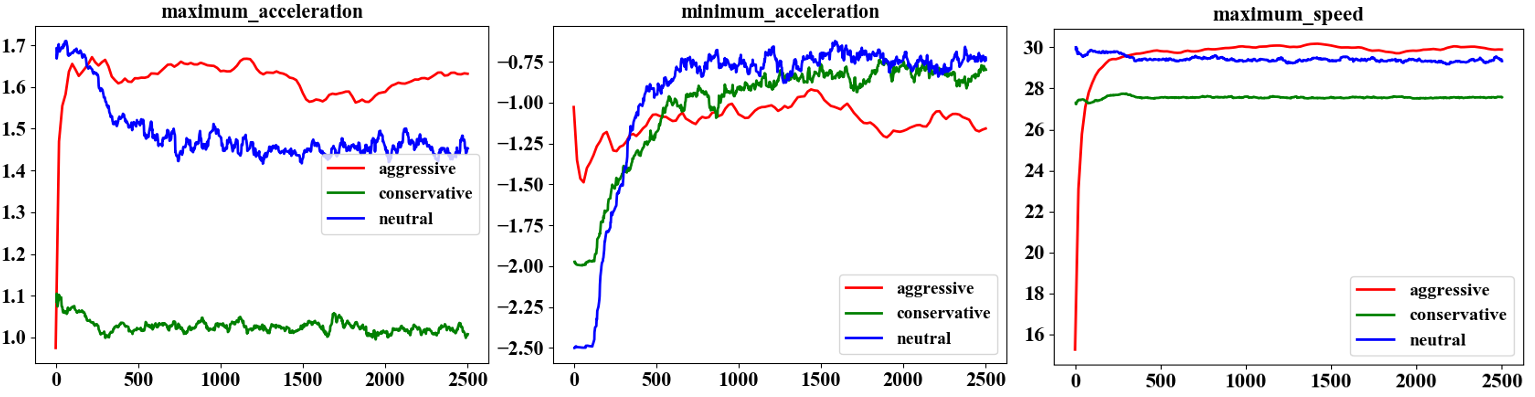}
        \setlength{\belowcaptionskip}{-10pt}
        \caption{\small Vehicle kinematic performance in maximum acceleration (left graph), minimum acceleration (middle graph), and maximum speed (right graph). The distinguishable curves in the three graphs indicate that the agent has learned different driving characteristics. Furthermore, the aggressive style (in red) shows higher values in maximum acceleration and maximum speed, and lower negative values in minimum acceleration, indicating that the meta-training procedure indeed encodes the features of different driving behaviors rather than just performing an average over them.}
        \label{fig:max_min}
    \end{figure*}
    
    \begin{figure*}[h]
        \centering
        \includegraphics[width=0.45\linewidth, scale=0.5]{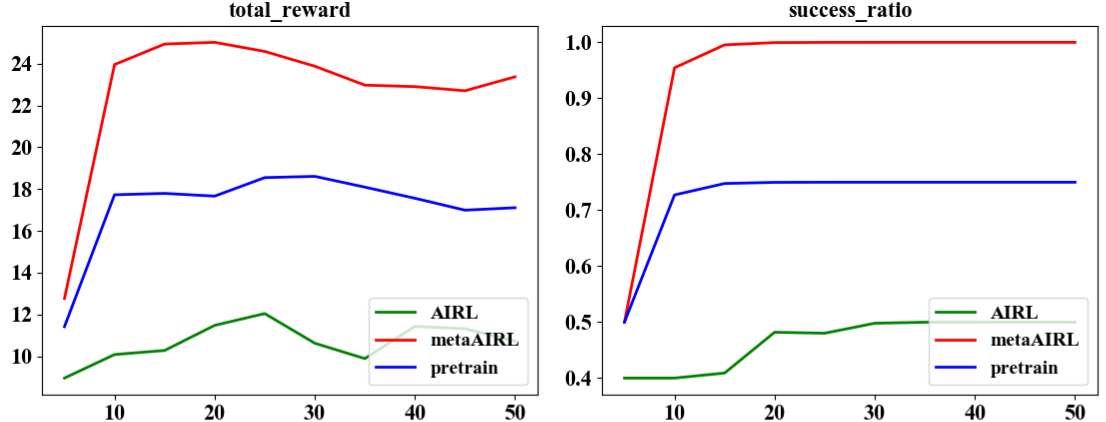}
        \setlength{\belowcaptionskip}{-10pt}
        \caption{\small Testing performance of three learning models. Our Meta-AIRL model (in red) shows higher rewards and episode success ratio than the other two baselines (the pretrained model in blue and the learning-from-scratch AIRL model in green), along with the increase of available samples (x-axis). It achieves the satisfactory generalization performance with only about 10 demonstrations from the meta-testing task, indicating the effectiveness of the fast adaptation of the proposed model.}
        \label{fig:new_test_curves}
    \end{figure*}

    \begin{figure*}[h]
        \centering
        \includegraphics[width=0.9\linewidth]{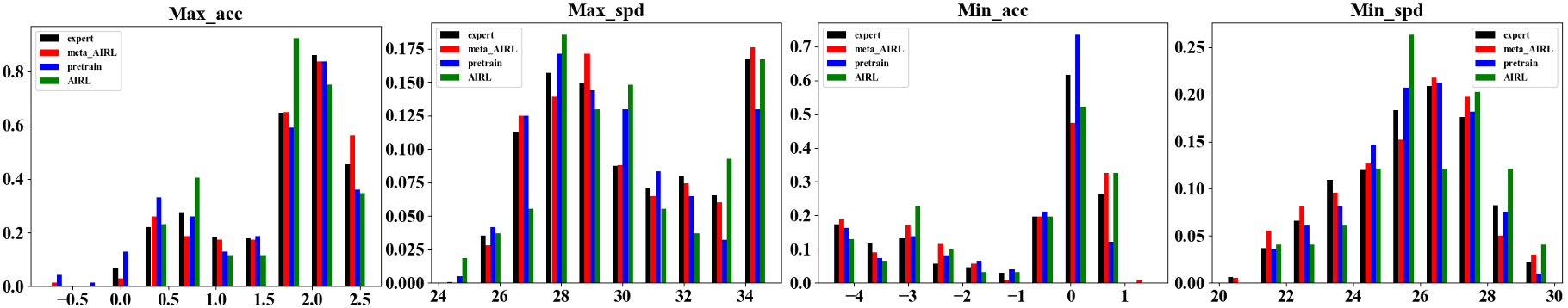}
        \setlength{\belowcaptionskip}{-10pt}
        \caption{\small Histograms of four driving kinematics (maximum acceleration, maximum speed, minimum acceleration, and minimum speed) in testing for three learned models and the expert. The distributions of the four metrics in the Meta-AIRL model (red bars) are much closer to the expert distributions (black bars) than those of the other two models (blue and green bars).}
        \label{fig:histogram}
    \end{figure*}

\subsection{Evaluation Metrics}
\label{subsec: metrics}
    
    We assess the effectiveness of the proposed method in four different aspects: the discriminator performance, the generator performance, the driving performance, and the vehicle kinematic performance. Within each aspect, we specify some metrics in detail. 
    
    \textbf{Discriminator Performance}. 
    The discriminator performance indicates the capability of distinguishing the expert data from the generated data. For a good discriminator, the value of $D_{\omega}(s_E,a_E)$, where $(s_E,a_E)$ is from expert dataset $\mathcal{S}^E$, should be as high as $1$ whereas the value of $D_{\omega}(s_G,a_G)$, where $(s_G, a_G)$ is from $\mathcal{S}^G$, should be as low as $0$ (Remember that we label expert data as $1$ and generated data as $0$). When trained with the generator, at optimal, the discriminator should output $0.5$ for both $D_{\omega}(s_E,a_E)$ and $D_{\omega}(s_G,a_G)$. 
    
    \textbf{Generator Performance}.
    The generator's capability is illustrated by the accumulated rewards in the roll-out episodes calculated by Equation (\ref{reward}). The larger the accumulated value is, the better the performance is. 
    
    \textbf{Driving Performance}.
    We use three metrics to evaluate the driving performance. Specifically, the roll-out step represents the duration of an episode before its termination (e.g. a crash, a prolonged driving, or a successful lane change). The decision-making step measures the duration of the decision-making process in an episode (e.g. how much time it uses to make the final decision). The episode success rate is the successful episodes over the total number of experimenting episodes. The value of the first two metrics should stay steady when the agent has learned the lane-change behavior, and the success rate should be as high as $1$ at optimal. 
    
    \textbf{Vehicle Kinematic Performance}. 
     To further explore the features of different driving styles, we evaluate some vehicle kinematics such as the maximum acceleration, the minimum acceleration, and the maximum speed, with the consideration that these metrics are representative to distinguish different habits. Ideally, they should be in different but stable ranges for different driving styles.

\subsection{Results}

    In this subsection, we show the meta-training results in Figure \ref{fig:d_rwd}, \ref{fig:steps_ratio}, and \ref{fig:max_min}, and the meta-testing results in Figure \ref{fig:new_test_curves} and \ref{fig:histogram}. In meta-training, we plot the performance curves of the two meta-training tasks as well as the online-testing curve of the meta-testing task adapted upon the current model parameters. In meta-testing, we compare the generalization performance of our model with two baselines. One is a pretrained model that is trained with all training tasks and then fine-tuned on the meta-testing task. This comparison is to show that the model from the meta-training phase does encode the patterns of different training tasks instead of just being integrated as an averaged model. The other baseline is a learning-from-scratch model that is trained based on AIRL on the meta-testing task with randomly initialized parameters. This comparison is to show how fast the our Meta-AIRL model can adapt to a satisfactory level.

\subsubsection{Meta-training Results}
    
    In Figure~\ref{fig:d_rwd}, the left and middle graphs show the probability curves of the discriminator. As mentioned earlier, a well-trained discriminator should output high probability values for expert data in its early stage as we use a high update frequency for it, and then goes down to 0.5 at optimal due to the adversarial effect from the generator, vice versa for the probability of the generated data. These two graphs in Figure \ref{fig:d_rwd} satisfactorily show this trend. 
    
    The generator's performance is reflected by the total reward curve in the right graph in Figure~\ref{fig:d_rwd}. The increasing and plateauing trend of the total rewards indicates the improved learning ability on the driving task.
    
    Figure~\ref{fig:steps_ratio} shows the driving performance which is represented by three metrics. The rolling-out steps in the left graph and decision-making steps in the middle graph show that the agent takes a long time to make lane changes in early episodes, but leans to make stable performance with fewer steps after trained for hundreds of episodes. The distinguishable curves in these two graphs represent different styles, indicating that the meta-training procedure does encode the characteristics of different tasks instead of just simply averaging over tasks. Furthermore, the online testing curves in red show that the task execution duration for aggressive drivers is shorter than that of conservative and neutral style drivers which is in line with our expectation. 
    
    The episode success ratio of the three driving styles is shown in the right graph in Figure \ref{fig:steps_ratio}. The curves increase quickly from low values to high ones and are quite close to 1, indicating that the trained agent can always make successful and acceptable lane-change decisions when requested.
    
    Figure~\ref{fig:max_min} displays the vehicle kinematic performance of the driving styles in three metrics, i.e. the maximum acceleration, minimum acceleration, and maximum speed. The distinguishable curves indicate that the agent does learn different driving characteristics of different styles, with different value ranges for their corresponding styles. More importantly, the aggressive style shows higher values in maximum acceleration and maximum speed, and lower values (in negative) in minimum acceleration. This further verifies that the trained model indeed encodes the features of different driving behaviors instead of just performing an average over them.

\subsubsection{Meta-testing Results}
    
    In meta-testing, we fine-tune the meta-trained model on the meta-testing task with different numbers of limited demonstrations, i.e. $\{5,10,15,...,50 \}$, separately, to check how well the model would be adapted when it has different data accessibility. The baseline models are also fine-tuned (for the pretrained baseline model) or trained (for the learning-from-scratch (AIRL) model) with the same numbers of demonstrations. The performance of the model is evaluated with 300 episodes rolled out in the meta-testing environment. The comparisons are shown in Figure~\ref{fig:new_test_curves} and \ref{fig:histogram}.
    
    Figure~\ref{fig:new_test_curves} displays the total rewards (left graph) and the success ratios (right graph) of the three learning methods. It is easy to observe that the Meta-AIRL model (red curves) performs much better than the other two baselines (blue and green curves): its total reward curve goes up much faster and stays higher, and its episode success ratio curve also rises up faster and is closer to 1. Such a good performance is achieved with only about 10 expert demonstrations through adaptation. 
    
    Figure~\ref{fig:histogram} shows the histograms of four driving kinematics, i.e. maximum acceleration, maximum speed, minimum acceleration, and minimum speed, of the expert data and the testing results of the three learning methods. As the number of test cases (300 tests) of the three trained models is different from that of the expert (3000 demonstrations), to make a fair comparison, we divide the bin's raw count by its corresponding total number of counts and the bin width, which is marked as y-axis. X-axis represents the metric values. The graphs visually show that the distributions of all of the four metrics in the Meta-AIRL model (red blocks) are much closer to the expert distributions (black blocks) when compared with the other two baselines (green and blue blocks). 
    
    The quantitative difference between the histogram from each learnt model and that from the expert is summarized in Table ~\ref{tab:table_example}. The deviation is measured by $l_1$ distance and calculated with values on y-axis. We can observe that Meta-AIRL has lower deviation values which confirms that Meta-AIRL compares favorably with the other two. Therefore, our Meta-AIRL model can faithfully recover the expert behavior with limited number of samples  for challenging novel tasks.
    
    
\begin{table}[h]
\caption{\small Comparison of three approaches in terms of the deviation of the learned distribution from the expert distribution. The deviation is measured by the $l_1$ distance between the histograms.}
\label{tab:table_example}
\begin{center}
\begin{tabular}{|c||c|c|c|}
\hline
	\textbf{Metrics} & \textbf{Meta-AIRL} &	\textbf{Pretrain model}	
	& \textbf{AIRL model} \\
\hline
\textbf{max\_acce}  &	 \bf{0.32} &	0.48 &	0.83 \\
			\hline
\textbf{max\_speed} &	\bf{0.08} &	0.18 &	0.27 \\
	\hline	
\textbf{min\_acce} &	\bf{0.30} & 0.41	& 0.42 \\
		\hline	
\textbf{min\_speed} &	\bf{0.15} &	0.28	& 0.33 \\
\hline
\end{tabular}
\end{center}
\end{table}

\section{Conclusion}
\label{sec:conclusion}

	In this work, we propose a framework (Meta-AIRL) to integrate Meta-learning and Adversarial Inverse Reinforcement Learning for fast adaptation of the learned model to new tasks. The model learns both the policy and reward function simultaneously from demonstrations by invoking a discriminator and a generator, and adapts quickly to new tasks with limited data samples by using different update frequencies and meta-learning rates for the discriminator and the generator. 
	
	The proposed model has been applied to the challenging decision-making task in the autonomous driving domain where the learning vehicle intensively interacts with surrounding agents. Meta-training results show that the agent indeed learns the inherent features of different driving styles instead of just performing an average over tasks. The comparison of Meta-AIRL with other baselines in the meta-testing results show that our model can fast adapt to new tasks with limited demonstrations and achieve satisfactory results comparable to those of the experts.

\bibliographystyle{IEEEtran}
\bibliography{IEEEabrv,example}

\end{document}